\documentclass{article}

\usepackage[nonatbib, final]{neurips_2022}

\usepackage[utf8]{inputenc} 
\usepackage[T1]{fontenc}    
\usepackage{hyperref}       
\usepackage{url}            
\usepackage{booktabs}       
\usepackage{amsfonts}       
\usepackage{nicefrac}       
\usepackage{microtype}      
\usepackage{graphicx}
\usepackage{multirow}
\usepackage{array}
\usepackage{xcolor} 
\usepackage{color, colortbl}
\usepackage{amsmath}
\usepackage{wrapfig}
\usepackage{caption}
\usepackage{algorithm}
\usepackage{listings}
\usepackage{arydshln}
\usepackage{bm}
\usepackage{booktabs}
\usepackage{lipsum}

\hypersetup{
  colorlinks,
  citecolor=violet,
  linkcolor=red
}

\newcommand{\ie}{\textit{i}.\textit{e}.}
\newcommand{\eg}{\textit{e}.\textit{g}.}
\newcommand{\etal}{\textit{et al}. }

\newlength\savewidth

 \makeatletter
\def\@fnsymbol#1{\ensuremath{\ifcase#1\or \dagger\or \ddagger\or
   \mathsection\or \mathparagraph\or \|\or **\or \dagger\dagger
   \or \ddagger\ddagger \else\@ctrerr\fi}}
\makeatother

\title{A Survey on Efficient Training of Transformers}

\author{Bohan Zhuang$^1$\thanks{Correspondence should be addressed to BZ. Email: $\tt  bohan.zhuang@gmail.com$} \quad Jing Liu$^1$ \quad Zizheng Pan$^1$  \quad Haoyu He$^1$ \quad Yuetian Weng$^1$ \quad Chunhua Shen$^2$ \\[0.2cm]
\centering{
	$^1$ZIP Lab, Monash University  ~~ ~
	$^2$Zhejiang University ~~ ~}}

\begin{document}

\maketitle

\begin{abstract}
Recent advances in Transformers have come with a huge requirement on computing resources, highlighting the importance of developing efficient training techniques to make Transformer training faster, at lower cost, and to higher accuracy 
by the efficient use of computation and memory resources.
This survey provides the first systematic overview of the efficient training of Transformers, covering the recent progress in acceleration arithmetic and hardware, with a focus on the former. We analyze and compare methods that save computation and memory costs for intermediate tensors during training, together with techniques on hardware/algorithm co-design. We finally discuss challenges and promising areas for future research.
\end{abstract}

\section{Introduction}

Deep learning is a recent most profound approach which has revolutionised machine learning (ML) and artificial intelligence and is leading the fourth industrial revolution. At its core, the success of deep learning depends on the vast computational resources available and an extremely large amounts of labeled data. Despite the huge excitement generated by the recent developments, deep learning models, especially Transformers \cite{vaswani2017attention}, have become formidably large and computationally intensive, resulting in two pressing challenges at the fundamental level.

The first issue concerns the intensive computation of training large Transformer-based models. A widely discussed energy study of deep learning models \cite{strubell2019energy} estimates that training a Transformer base model with neural architecture search (NAS) \cite{so2019evolved} produces about 626,155 pounds of planet-warming carbon dioxide, equal to the lifetime emissions of five cars; as models grow bigger, their demand for computing is outpacing improvements in hardware efficiency. For example, GPT-3 \cite{brown2020language} (the precursor to
ChatGPT) was trained on half a trillion words and equips with 175 billion parameters.
Notably, according to the technical overview of GPT-3\footnote{\url{https://lambdalabs.com/blog/demystifying-gpt-3/}}, it would take 355 GPU-years and cost at least \$4.6M for a single training run, estimated with theoretical 28 TFLOPS for V100 and lowest 3-year reserved cloud pricing. 
Therefore, the true
groundbreaking success of deep learning, such as ChatGPT, is exclusively dominated by large and rich enterprises such
as Google or Microsoft.
It becomes extremely important to make
deep learning tenable in computation and energy efficiency for Green AI \cite{schwartz2020green}, and democratize AI to wider communities with limited resources.

The second issue comes with the exponentially growing training memory proportional to the attention-based model size. For example, the largest language model in literature grows from 345M with BERT-large \cite{kenton2019bert} in 2018, to hundreds of billions till now with models such as MT-NLG \cite{smith2022using} equipped with 530B parameters. 
Therefore, these SOTA massive models call for memory efficient training techniques to reduce the memory footprint of storing intermediate tensors and data exchanges (communications) across accelerators, while ensuring high processing elements (PE) utilization.

\begin{figure*}[t!]
    \centering
    \includegraphics[scale=0.5]{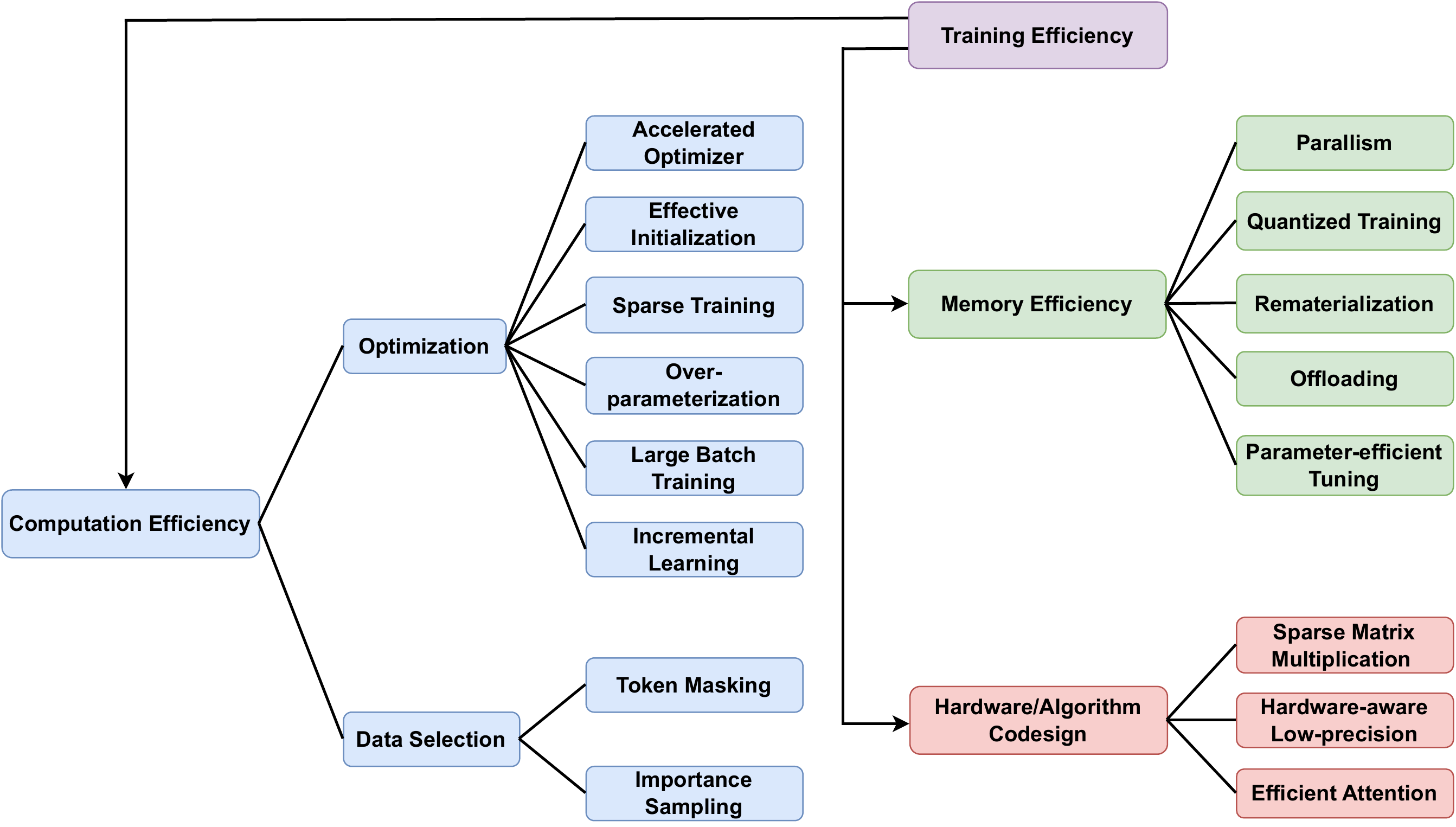}
    \caption{Overview of the efficient training of Transformers including computation efficiency, memory efficiency and hardware/algorithm codesign perspectives.}
    \label{fig:main}
    \vspace{-1em}
\end{figure*}

In this survey, we review the generic techniques that boost computation and memory efficiency for training attention-based models, \ie, Transformers, as shown in Figure \ref{fig:main}. 
We characterize them by the technical innovations and primary use case, summarize them and draw connections between them. 
We are primarily interested in arithmetic innovations that improve the training efficiency of Transformers and also briefly discuss hardware/algorithm codesign advances. We leave the review of hardware accelerator design, a broad class, as future work.

\section{Computation Efficiency}
\label{sec:computation}

\subsection{Optimization}

\noindent \textbf{Optimizer}. 
To achieve a faster convergence rate for gradient descent, a classic solution is to fuse the momentum technique, where each step is a combination of the steepest descent direction and the most recent iterate displacement, helping to accelerate gradient descent in the relevant direction and dampens oscillations.
The seminal works include Nesterov's accelerated gradient \cite{nesterov1983method} for convex optimization and proximal gradient with momentum \cite{li2017convergence} towards non-convex problems, etc. 
To meet the demand of large-scale optimization of machine learning models, dominant optimizers are designed in a stochastic fashion. In particular, stochastic gradient descent (SGD) with momentum
and the adaptive learning rate estimation method Adam \cite{kingma2015adam} are widely used to train deep neural networks. Empirically, training Transformers with Adam outperforms the SGD counterpart, and \cite{zhang2019adam} demystifies that a heavy-tailed distribution of the noise in stochastic gradients is the main cause of SGD’s poor performance and understands Adam through the lens of adaptive noise clipping. 
By default, AdamW \cite{loshchilov2019decoupled}, a variant of Adam which decouples the $L_2$ regularization and the weight decay, is the most widely used optimizer for Transformers. More recently, Google searches optimization algorithms and discovers a simple and effective
optimizer called Lion \cite{chen2023symbolic}. Lion only
keeps track of the momentum with the first-order gradient, and 
its update only considers the $\rm{sign}$ direction and has the same magnitude for
each parameter, which is very different from the adaptive optimizers like AdamW. In practice, Lion in general converges faster, and is more memory-efficient and accurate than AdamW for training Transformers on various benchmarks. 
We refer readers to \cite{lin2020accelerated,bottou2018optimization} for more details about accelerated optimization methods in machine learning.

To improve the generalization of Transformers, Sharpness-aware minimization (SAM) \cite{foret2021sharpness} seeks to simultaneously minimize loss value and loss sharpness, based on 
the connection between the geometry of the loss landscape and generalization, \ie, a flatter minimum tends to improve generalization. 
The following work \cite{chen2021vision} applies SAM to Transformer, observing significant accuracy gains via smoothing the loss surface. 
However, SAM needs to solve a bi-level min-max optimization problem, which nearly doubles the training time. To accelerate optimization, \cite{du2021efficient} proposes stochastic weight perturbation to preserve the generalization capability and sharpness-sensitive subset selection strategies. More recently, \cite{du2022sharpness} designs a near zero-cost proxy of the sharpness loss by replacing the sharpness estimation as the KL-divergence between the two consecutive update steps.

\noindent \textbf{Initialization}. 
A good initialization is essential to stabilize training, enable higher learning rate, accelerate convergence, and improve generalization. Thus, many works have been proposed for better initialization of Transformers. Specifically, Fixup \cite{zhang2019fixup} proposes to properly rescale a standard initialization to ensure proper gradient norm to avoid exploding or vanishing gradients, which can train very deep networks with over 10,000 layers without adding normalization layers. 
Based on the insight that the function computed by normalized residual blocks is close to the identity function (\ie, unit variance), the following works ReZero \cite{bachlechner2021rezero} and SkipInit \cite{de2020batch} 
simply initialize each layer to perform the identity operation. Specifically, they add a learnable scaling multiplier on the output of each residual block: 
\begin{equation}
   {\bf{x}}_{l+1} = {\bf{x}}_l + \alpha_lF_l({\bf{x}}_l),
\end{equation}
where ${\bf{x}}_l$ and $F_l(\cdot)$ are the input and the function at layer $l$, where the function can be multi-head self-attention layers (MSA) and feed-forward networks (FFN), and $\alpha_l$ is simply initialized to 0.
Customized to Transformers, T-Fixup \cite{huang2020improving} analyzes that part of the optimization difficulty comes from the unstable early updates in the Adam optimizer as the variance of the second-order momentum is unbounded. Therefore, it follows Fixup to adopt rescaling schemes for the initialization of residual blocks.
All the above-mentioned methods remove batch/layer normalization from all blocks and train without learning rate warmup.
On training deep vision Transformers (ViT), \cite{touvron2021going} proposes channel-wise learnable scaling factors and empirically observe that re-introducing the warmup and layer normalization techniques can make training more stable. 

Apart from the rescaling paradigm, some literature proposes to improve initialization from a new perspective by leveraging the relationship between self-attention and convolutional layers. 
\cite{cordonnier2020relationship} proves that a multi-head self-attention layer with $N$ heads and a relative positional encoding with dimension $D\geq3$ can be reparameterized to express any convolutional layer of filter size $\sqrt{N} \times \sqrt{N}$. The attention can be decomposed into a content term and a (relative) positional term, where the latter determines the center and width of attention of each head.
Based on this property, ConViT \cite{d2021convit} learns to control the locality by adding a soft gating parameter to balance the two terms, which has the effect of incorporating soft convolutional inductive biases into global self-attention.

\noindent \textbf{Sparse training}.
The key idea of sparse training is to directly train sparse subnetworks instead of the full networks from scratch without sacrificing accuracy. The reliability was first demonstrated by the lottery ticket hypothesis (LTH) \cite{frankle2019lottery} that a dense, randomly initialized network contains subnetworks (winning tickets) which can be trained in isolation to match the accuracy of the original network. However, LTH requires identifying the winning tickets in an alternating train-prune-retrain manner, which makes the training extremely costly for large models and datasets, limiting the practical benefits. In light of this, follow-up works with higher training efficiency can be roughly categorized into three categories: (i) 
find sparse networks once at initialization by measuring the importance of connections on the loss, eliminating the need for the complex iterative optimization schedule \cite{lee2019snip,wang2020picking} ; (ii) identify the winning tickets in Transformers 
at a very early training stage via low-cost schemes and then merely train these early tickets until convergence \cite{you2019drawing,chen2021earlybert}; (iii) 
use an alternating pruning and growing schedule to dynamically update model sparsity patterns throughout training, suitable for general architectures \cite{evci2020rigging,chen2021chasing}.

\noindent \textbf{Overparameterization}.
Practical DNNs are heavily overparameterized, where the number of learnable parameters is much larger than the number of training samples. It is observed that overparameterization empirically improves both convergence and generalization, with theoretical guarantee though not sufficient. The early work \cite{arora2018optimization} mathematically proves that increasing depth as overparameterization in linear neural networks can accelerate SGD convergence. \cite{li2018learning} further explores two-layer non-linear neural networks and \cite{allen2019convergence} proves that SGD can converge to global minima on the training objective of DNNs in \textit{polynomial time}, assuming training samples are not duplicated, and the number of parameters is polynomial in the number of training samples and network depth. 
In terms of generalization, \cite{allen2019learning} theoretically proves that a sufficiently overparameterized (three-layer) neural network generalizes to the population risk and an intriguing property is that \textit{there exists an accurate network in the close neighborhood of any point on the SGD training trajectory with high probability over random initialization.} Note that it has deep connections with LTH as it partially explains why LTH stands in sparse training as good small sub-networks with low risks are plentiful due to overparameterization. Applied to Transformers, \cite{li2020train} exploits the faster convergence and better generalization from the overparameterization theory to design an efficient training pipeline: training a very large model, then perform early stopping and heavily compress it, analogous to LTH.

\noindent\textbf{Large batch training}. 
Another prevailing way to accelerate training is to use a large batch size, delivering a reduced number of iterations per epoch and better computing resource utilization. From the statistical view, large batch training reduces the variance of the stochastic gradient estimates, so a reliable step size needs to be tuned for better convergence \cite{bottou2018optimization}. At the era of convolutional neural networks, \cite{goyal2017accurate} employs the 
linear scaling of the learning rate to train ResNet-50 on ImageNet with a batch size of 8,192 in 1 hour. More advanced step size estimation methods are then proposed. The widely used methods are LARS \cite{you2017large} for SGD and LAMB \cite{you2019large} for Adam, which propose to use layerwise adaptive learning rates for ResNet and Transformers respectively.
The layerwise adaptation strategy can be formulated as 
\begin{equation}
 {\bf{w}}_{t+1}^i = {\bf{w}}_t^i - \eta_t\frac{\phi(\left\| {\bf{w}}_t^i \right\|)}{\left\| \bm{\gamma}_t^i \right\|}\bm{\gamma}_t^i,   
\end{equation}
where $\eta_t$, ${\bf{w}}_t^i$ and $\bm{\gamma}_t^i$ are the learning rate, parameters and the momentum-based gradients of the $i$-th layer at time step $t$, $\phi$ is a scaling function.
It equips with a normalization term that provides robustness to exploding gradients and plateaus, and the scaling term ensures that the norm of the update is of the same order as that of the parameter, promoting faster convergence.
More recently, more powerful optimization methods customized to large batch training have been empirically shown to perform well. For example, \cite{kaddour2022stop} shows that averaging the weights of a certain number of latest checkpoints can facilitate faster training. DeepMind in \cite{hoffmann2022training} trains over 400 Transformer language models with varying scales of model size and \# of training tokens, reaching to a practical hypothesis that the model size and the number of training tokens should be scaled equally for compute-optimal LLM training.

\noindent \textbf{Incremental learning}. 
The high-level concept of incremental learning is relaxing the original challenging optimization problem into a sequence of easy-to-optimize sub-problems, where the solution of one sub-problem can serve as a good initialization to the subsequent one to circumvent the training difficulty, in analogy with annealing. Some works \cite{gong2019efficient,gu2021transformer} propose to accelerate BERT pretraining by progressively stacking layers, properly initializing a larger model from a smaller one. \cite{zhang2020accelerating} goes in a reverse direction to train Transformers with stochastic depth via layer dropping, where it progressively increases dropping rate along both time dimension and depth dimension. Customized to ViT, AutoProg \cite{li2022automated} proposes to automatically decide whether, where and how much should the model grow during progressive learning using neural architecture search. A key observation is that progressively increasing the resolution of the input images (reducing the patch size) can significantly accelerate ViT training, aligning with the widely known training dynamics that focus on low-frequency structure in the early stage and high-frequency semantics in the latter stage.

\subsection{Data Selection}

Apart from the model efficiency, data efficiency is also a crucial factor of efficient training. 

\noindent \textbf{Token masking}. Token masking is a dominant approach in self-supervised pre-training tasks, such as masked language modeling (MLM)~\cite{kenton2019bert,brown2020language} and masked image modeling (MIM)~\cite{bao2022beit,he2022masked}. The spirit of token masking is to randomly mask some input tokens and train the model to predict the 
missing content, e.g., vocabulary id or pixels, with the context information from the visible tokens. Since squeezing the sequence length reduces both the computational and memory complexity quadratically, skipping processing the masked tokens brings considerable training efficiency gain for MLM and MIM. For MLM, \cite{song2019mass} proposes to jointly pre-train the encoder and decoder for language generation tasks while removing the masked tokens in the decoder to save memory and computation costs. For MIM, representative work~\cite{he2022masked} shows that in vision, removing the masked image patches before the encoder demonstrates stronger performance and 3$\times$ or more lower overall pre-training time and memory consumption than keeping the masked tokens. A similar phenomenon is also found in~\cite{li2022scaling} that for language-image pre-training, randomly masking and removing the masked image patches shows 3.7$\times$ faster overall pre-training time
than the original CLIP~\cite{radford2021learning}.

\noindent\textbf{Importance sampling}.
Importance sampling over data, also known as data pruning, is theoretically guaranteed to accelerate stochastic gradient algorithms for supervised learning by prioritizing informative training examples, mainly benefiting from variance reduction. For DNNs, a principal way of estimating per-sample importance is to use gradient norm, and \cite{katharopoulos2018not,johnson2018training} use different approximations to make calculating these norms tractable. \cite{paul2021deep} further speeds up the sampling process similar to the early-bird LTH, but in the data domain, that simple average gradient norms or error $\ell_2$-norms over several weight initializations can be used to identify important examples at the very early stage in training. More recently, \cite{sorscher2022beyond} shows an exciting analytic theory that \textit{the scaling of test error with dataset size can break beyond power scaling laws and be reduced to at least exponential scaling if equipped with a superior data pruning metric}, and it employs a self-supervised metric using $k$-means clustering. It demonstrates a promising direction towards more efficient neural scaling laws based on data importance sampling.

\begin{table}[htp]
\centering
\renewcommand\arraystretch{1.2}
\caption{Summary of memory efficient training methods. Abbreviations include: AMP= Automatic Mixed Precision, DP = Data Parallelism, MP = Model Parallelism, TP = Tensor Parallelism, PP = Pipeline Parallelism, ACT = Activation Compressed Training and PET = Parameter-efficient Tuning. }
\label{tab:memory}
\scalebox{0.9}{
\begin{tabular}{c|c}
Method & Class  \\ \hline
Micikevicius \etal \cite{low_pre_train}  & AMP \\
Chen \etal \cite{chen2016training} & Rematerialization \\
Herrmann \etal \cite{herrmann2019optimal} & Rematerialization \\
ZeRO-Offload \cite{ren2021zero}& Offloading \\
Beaumont \etal\cite{beaumont2021efficient} & Offloading + Rematerization \\\hdashline
ZeRO \cite{rajbhandari2020zero} & DP+MP+AMP  \\
Megatron-LM \cite{shoeybi2019megatron} & DP+TP   \\
GPipe \cite{huang2019gpipe} &DP+PP  \\
torchgpipe \cite{kim2020torchgpipe} &PP+Rematerization\\
Megatron-LM$^{*}$\cite{narayanan2021efficient} &DP+TP+PP+AMP \\\hdashline
Wang \etal \cite{wang2018training} & FP8 Training \\
Cambier \etal \cite{cambier2020shifted} & FP8 Training \\
Mesa \cite{pan2021mesa} & 8-bit ACT \\
ACTNN \cite{actnn}, GACT \cite{gact}  & 2-bit ACT  \\\hdashline
\cite{lester2021power, jia2022vpt, houlsby2019parameter}  & Addition-based PET \\\hdashline
Bitfit \cite{zaken2022bitfit}, LoRA \cite{hu2022lora}  & Reparameterization-based PET \\
\end{tabular}
}
\end{table}
\vspace{-1em}

\section{Memory Efficiency}

Apart from the computation burden, the growing model size of large Transformer models, \eg, from BERT \cite{kenton2019bert} 345M parameter model to GPT-3 of 1.75 trillion parameters, is a key bottleneck for training as they do not fit into the memory of a single device. 
We first analyze the memory consumption of the existing model training frameworks, which is occupied by 1) \textit{model states}, including optimizer states (\eg, momentum and variance in Adam), gradients and parameters; and 2) \textit{activations} (we ignore temporary buffers and idle fragmented memory as they are relatively small). 
We summarize the memory efficient training methods in Table~\ref{tab:memory}.
In the following, we discuss dominant solutions to optimize memory usage. 

\noindent \textbf{Parallelism}. 
Training large DNNs with parallelism across devices is a common practice to meet the memory demands. There are basically two paradigms: \textit{Data Parallelism} (DP) which distributes a minibatch of data across different devices and \textit{Model Parallelism} (MP) which allocates subgraphs of a model across multiple workers. For DP, with the increase of available workers, the batch size is close to linear scaling.
Large batch training discussed in Sec. \ref{sec:computation} is developed for this case. 
However, it is obvious that DP has high communication/computation efficiency but poor memory efficiency - when model becomes large, the single device cannot store the model replica and the synchronized communications for gradients can hinder the scalability of DP. Therefore, DP itself is only suitable for training small to moderate models. To improve the scalability of DP, one solution for Transformer is parameter sharing \cite{lan2019albert}, known as Albert, but it limits the representational power. More recently, ZeRO \cite{rajbhandari2020zero} incorporates uniform partitioning strategy with DP, where each data parallel process merely deals with one partition of model states, working in a mixed precision regime. 
To deal with very large DNNs, one always need to utilize model parallelism to allocate different layers across multiple accelerators in a ``vertical'' manner. Though MP has good memory efficiency, its communication and computation efficiency is low due to high volume data transfer cross devices and poor PE utilization. Luckily, there are two strategies to further boost MP efficiency in an orthogonal ``horizontal'' dimension, including \textit{Tensor Parallelism} (TP) and \textit{Pipeline Parallelism} (PP). 
TP partitions a tensor operation in a layer across workers for faster computation and more memory saving. Customized to Transformer-based models, Megatron-LM \cite{shoeybi2019megatron} slices both MSA and FFN across GPUs and requires only a few extra All-Reduce operations in the forward and backward pass, allowing them to train models up to 8.3 billion parameters using 512 GPUs.
In terms of PP, it was originally proposed in GPipe \cite{huang2019gpipe}, which splits the input mini-batch into multiple smaller micro-batches, enabling different accelerators (sequential layers are partitioned across accelerators) to work on different micro-batches simultaneously before applying a single synchronous gradient update for the entire mini-batch. However, it still suffers from pipeline bubbles (accelerator idle time) that reduce efficiency. In particular, PyTorch implements the torchgpipe \cite{kim2020torchgpipe}, which performs micro-batch PP with checkpointing, allowing scaling to a large number of micro-batches to minimize the bubble overhead.

\noindent Note that DP and MP are orthogonal and so one can use both simultaneously to train larger models with higher computation and memory capacity. For example, Megatron-LM$^{*}$ \cite{narayanan2021efficient} and DeepSpeed \cite{rasley2020deepspeed}
compose tensor, pipeline, and data parallelism to scale training to thousands of GPUs.

\noindent \textbf{Quantized training}.
\label{sec:quantized_training}
The standard routine for training neural networks adopts full-precision (\ie, FP32). In contrast, quantized training trains neural networks from scratch in reduced precision by compressing the activations/weights/gradients into low-bit values (\eg, FP16 or INT8).
It has been shown in previous works that reduced precision training \cite{dorefa,HubaraCSEB17} can accelerate neural network training with favorable performance. 
For Transformers, the most widely adopted approach is automatic mixed-precision (AMP) training~\cite{low_pre_train}. Specifically, AMP stores a master copy of weights in full-precision for updates while the activations, gradients and weights are stored in FP16 for arithmetic. Compared to full-precision training, AMP is able to achieve faster training/inference speed and reduce memory consumption during network training. For example, based on a batch size of 64 and image resolution of $224\times224$, training a DeiT-B~\cite{deit} on RTX3090 under AMP is 2$\times$ faster than full-precision training (305 \textit{vs.} 124 images/s), as well as consuming 22\% less peak GPU memory (7.9GB \textit{vs.} 10.2GB).
While it is commonly believed that at least 16-bits is necessary to train networks without impacting model accuracy~\cite{low_pre_train,0002MMKAB0VKGHD18}, the most recent support for FP8 training on NVIDIA H100 has shown promising results on Transformer training, where training DeiT-S and GPT~\cite{brown2020language} under FP8 can match those of 16-bit training.
Apart from reduced precision training which simultaneously quantizes activations/weights/gradients, activation compressed training (ACT)~\cite{actnn} stores low-precision approximate copies of activations while computing the forward pass exactly, which helps to reduce the overall memory consumption during training. The saved activations are then dequantized to the original precision in the backward pass to calculate gradients. 
Recent work~\cite{pan2021mesa,gact} further propose to customize ACT to support memory-efficient Transformer training.

\noindent \textbf{Rematerialization and offloading}.
\textit{Rematerialization}, also known as \textit{checkpointing} \cite{chen2016training}, is a widely used technique for space-time tradeoff that only stores a portion of activations/weights during the forward pass and recomputes the rest during the backward pass. \cite{chen2016training} provides a simple periodic schedule which was implemented in PyTorch\footnote{\url{https://pytorch.org/}}, but it is only optimal for homogeneous sequential networks. More advanced methods such as \cite{herrmann2019optimal} implements optimal checkpointing for heterogeneous networks\footnote{\url{https://gitlab.inria.fr/hiepacs/rotor}}. 
In terms of \textit{offloading}, it is a technique to use external memory such as CPU memory as an extension of GPU memory to increase memory capacity during training, through communications between GPU and CPU. The model states as well as activations, can be offloaded to CPU, but the optimal choice needs to minimize communication cost (\ie, data movement) to/from GPU, reduce CPU computation and maximize GPU memory saving. 
A representative work is ZeRO-Offload \cite{ren2021zero}, which offers optimal offloading strategy customized to mixed-precision training with Adam optimizer. It offloads all fp32 model states and the fp16 gradients on the CPU memory, and computes the fp32 parameter updates on CPU. The fp16 parameters are kept on GPU and the forward and backward computations are on GPU. For the best of both worlds, \cite{beaumont2021efficient} proposes to jointly optimize activation offloading and rematerialization.

\noindent\textbf{Parameter-efficient tuning.}
The public model zoo represented by HuggingFace, which contains rich pretrained models that are ready to be downloaded and executed, is contributing significantly to reductions in training costs.
Efficient tuning these readily available models is becoming a prevailing way to drastically cut training costs.
As a powerful alternative for the vanilla full fine-tuning, parameter-efficient tuning (PET) only updates a small number of additional parameters while freezing the pretrained model to significantly reduce the storage burden, which scales with dynamic deployment scenarios without the need to store a separate instance of model for each case. The general PET approaches can be categorized into \textit{addition-based} methods and \textit{reparameterization-based} methods. The former attaches additional trainable parameters to the pretrained model and only tune these parameters. For example,
\cite{lester2021power,jia2022vpt}
add trainable parameters to the input space, and \cite{houlsby2019parameter} adds the adapter module twice to each Transformer block after the MSA and FFN. However, the extra parameters introduce additional computation and memory overhead during inference. To tackle this challenge, the latter proposes to tune parameters that are inherently in the model \cite{zaken2022bitfit} or new parameters that can be reparameterized into the model \cite{hu2022lora}, thereby yielding no sacrifice on the inference efficiency. Inspired by the observation that large language pretrained models have low intrinsic dimension \cite{aghajanyan2021intrinsic}, the representative work LoRA \cite{hu2022lora} approximates the update of self-attention weights into two low-rank matrices, which can be merged into the pretrained weights during inference.
Notably, one of the most recognized efforts for democratizing LLM is Stanford Alpaca \cite{alpaca}, which is fine-tuned from the open-sourced LLaMA models \cite{touvron2023llama} using the 52K instruction-following data generated from ChatGPT. To fine-tune it cheaply and efficiently, its variant Alpaca-LoRA \footnote{\url{https://github.com/tloen/alpaca-lora}} further adopts the low-rank LoRA to enable instruct-tuning LLaMA on customer hardware, showing training can be done within hours on a single RTX 4090.

\noindent\textbf{Open-source frameworks.}
There are several widely adopted prototypes for training large Transformer models at scale, in which Microsoft DeepSpeed\footnote{\url{https://github.com/microsoft/DeepSpeed}}, HPC-AI Tech Colossal-AI\footnote{\url{https://github.com/hpcaitech/ColossalAI}} and Nvidia Megatron-LM\footnote{\url{https://github.com/NVIDIA/Megatron-LM}} are the pioneering ones. Specifically, DeepSpeed is implemented mainly based on \cite{rasley2020deepspeed} and ZeRO series works \cite{rajbhandari2020zero,ren2021zero}, Colossal-AI is built upon \cite{bian2021colossal}, and Megatron-LM implements \cite{narayanan2021efficient}. All of these support data and model parallelism in mixed precision, along with other general practices such as offloading and rematerialization. More libraries for efficient distributed training include but not limited to HuggingFace Transformers\footnote{\url{https://github.com/huggingface/transformers}}, MosaicML Composer\footnote{\url{https://github.com/mosaicml/composer}}, Baidu PaddlePaddle\footnote{\url{https://github.com/PaddlePaddle/Paddle}}, Bytedance Lightseq\footnote{\url{https://github.com/bytedance/lightseq}}, EleutherAI GPT-NeoX\footnote{\url{https://github.com/EleutherAI/gpt-neox}}, etc. 

\vspace{-0.5em}
\section{Hardware/Algorithm Co-design}
Apart from computing and memory burden, designing efficient hardware accelerators enables faster training and inference for DNNs. Specifically, compared with central processing units (CPUs), graphics processing units (GPUs) are more powerful to perform matrix multiplication due to the high degree of parallelism. For applications that focus on specific computation tasks, application-specific integrated circuits (ASICs) have the advantage of low power consumption, and high training/inference speed. For example, a tensor processing unit (TPU) designed by Google delivered 30$\sim$80$\times$ higher performance-per-watt than contemporary CPUs and GPUs~\cite{jouppi2017datacenter}. However, ASICs are not easily reprogrammable or adaptable to a new task. In contrast, Field Programmable Gate Arrays (FPGAs) are designed to be reprogrammed to perform different functions as needed, and can also be used as a prototype for ASICs before finalizing the design. To further optimize the training efficiency of DNNs, especially Transformers, hardware-algorithm co-design takes the constraints and capabilities of the hardware into account when designing the algorithm, which will be introduced in the following subsections.

\noindent \textbf{Sparse matrix multiplication}. To reduce the computation overhead of Transformers, sparse general matrix multiplication (SpGEMM), which involves multiplying a sparse matrix with a dense matrix, takes advantage of the sparsity of the attention matrices to reduce the number of computations. There are several popular sparse matrix computation libraries, such as Intel Math Kernel Library\footnote{\url{https://www.intel.com/content/www/us/en/developer/tools/oneapi/onemkl.html}} on CPU and cuSPARSE\footnote{\url{https://docs.nvidia.com/cuda/cusparse/}}, CUSP\footnote{\url{https://cusplibrary.github.io/}} and 2:4 structured sparsity \footnote{\url{https://developer.nvidia.com/blog/accelerating-inference-with-sparsity-using-ampere-and-tensorrt/}}\cite{mishra2021accelerating} on GPU.
However, due to the irregular sparsity, SpGEMM is often hardware unfriendly to general-purpose processors, such as CPUs and GPUs. To handle this, specialized hardware accelerators, such as FPGAs and ASICs, are required to handle the poor data locality issue. For example, OuterSPACE~\cite{pal2018outerspace} transforms matrix multiplication into an outer product procedure and eliminates redundant memory accesses by decoupling multiplication from accumulation. To take full advantage of this reduction without introducing significant overheads, OuterSPACE builds a custom accelerator with reconfigurable memory hierarchy and achieves a mean speedup of 7.9$\times$ over the CPU running Intel Math Kernel Library and 14.0$\times$ against the GPU running CUSP. Furthermore, to alleviate the data movement bottleneck caused by high sparsity, ViTCoD~\cite{you2023vitcod} uses a learnable auto-encoder to compress the sparse attentions to a much more compact representation and designs encoder and decoder engines to boost the hardware utilization.

\noindent \textbf{Hardware-aware low-precision}. Lowering the precision of the computations reduces the amount of memory and computation, which can be implemented in hardware-friendly fixed-point or integer representations instead of floating-point ones. As a result, we can use lower precision multipliers, adders, and memory blocks, resulting in a significant improvement in power consumption and speedup. Moreover, low-precision arithmetic can be combined with other techniques, such as pruning and low-rank approximation, to achieve further acceleration. For example, Sanger~\cite{lu2021sanger} uses 4-bit queries and keys to compute the quantized prediction of sparse attention matrix. Then, the sparse attention masks are rearranged into structured blocks and handled by reconfigurable hardware. The following work DOTA~\cite{qu2022dota} identifies unimportant connections in attention using low-rank transformation and low-precision computation. By incorporating token-level parallelism and out-of-order execution, DOTA achieves a 152.6$\times$ speedup over GPU.

\noindent \textbf{Efficient attention.} Apart from the sparse matrix multiplication and low-precision computation, several pioneering works focus on efficient and lightweight attention implementation in hardware~\cite{ham20203,ham2021elsa,dao2022flashattention}. Specifically, A$^3$~\cite{ham20203} only selects those keys that are likely to have high similarity with the given queries to reduce the amount of computation in attention. 
ELSA~\cite{ham2021elsa} filters out irrelevant keys for a particular query based on the hashing similarity to save computation. 
With an efficient hardware accelerator, ELSA achieves a speedup of 58.1$\times$ as well as three orders of magnitude improvements in energy efficiency compared to an Nvidia V100 GPU equipped with 16GB memory.
Notably, FlashAttention~\cite{dao2022flashattention} proposes to exploit tiling to reduce the I/O communication between GPU high bandwidth memory (HBM) and on-chip SRAM, which is becoming a default fast and memory-efficient attention module for speedup.

\vspace{-0.1em}
\section{Conclusion and Future Research}

In this survey, we have reviewed several important factors that improve the training of Transformers: 1) appropriate initialization and optimization paradigms that can accelerate the convergence rate with fewer training iterations, resulting in \textit{lower computational costs}; 2) higher data efficiency by sampling informative training samples towards \textit{more efficient neural scaling laws} of test error with respect to dataset size; 3) memory-efficient techniques to meet the memory requirements for training large Transformers, which requires \textit{jointly optimizing PE utilization, memory and communication footprints across accelerators}, using parallelism, low-precision arithmetic, checkpointing and offloading strategies, etc;
4) \textit{hardware and algorithm co-design} to maximize the training scalability on hardware platforms.

We finally highlight several promising directions based on the reviewed progress: 
(i) joint training and deployment efficiency optimization. In reality, we usually need to deploy models to diverse tasks and platforms with different resource constraints. 
Therefore, it is highly desirable to develop new methods for efficiently training an elastic supertnet that supports many diverse architectural configurations following single-shot NAS \cite{chen2021autoformer} or mixture of experts \cite{fedus2022switch}, for multiple tasks and budget requirements;
(ii) on-device training \cite{lin2020mcunet} on the edge with limited resources (\eg, low battery and memory capacity), to avoid frequently transmitting data which results in privacy and latency issues; 
(iii) combine efficient approximation techniques such as token/model pruning, low-rank factorization, lightweight architecture design, dynamic neural networks and etc, to reduce the model size and computational cost in the complimentary sense;
(iv) a standard benchmark to evaluate and compare the efficient training methods. Having such a benchmark would fasten the adoption of these methods and lead to actual costs reduction down the line.

\bibliographystyle{abbrv}
{
 	\small
	\bibliography{reference}
}

\end{document}